\documentclass[runningheads]{llncs}
\usepackage[utf8]{inputenc}
\usepackage[english]{babel}
\usepackage{amsmath}
\usepackage{amsfonts}
\usepackage{amssymb}
\usepackage{color}
\usepackage{mathtools}
\usepackage{graphicx}

\usepackage{tikz}
\usetikzlibrary{fit,positioning,quotes,arrows.meta}
\usepackage{subcaption}
\pdfoutput=1

\DeclareMathOperator*{\argmax}{arg\,max}
\DeclareMathOperator{\E}{\mathbb{E}}
\DeclareMathOperator{\U}{\mathbf{U}}
\DeclareMathOperator{\DKL}{\text{D}_\text{KL}}
\DeclareMathOperator{\F}{\text{F}_{\text{par}}}

\begin{document}
\title{Bounded Rational Decision-Making with Adaptive Neural Network Priors}
\titlerunning{Bounded Rational Decision-Making with Adaptive ANN Priors}
%
\author{Heinke Hihn \and
Sebastian Gottwald \and
Daniel A. Braun}
%
\authorrunning{H. Hihn et al.}
%
\institute{Institute for Neural Information Processing \\ Faculty of Engineering, Computer Science, and Psychology \\ Ulm University, Ulm (Germany) \\
\email{\{heinke.hihn, sebastian.gottwald, daniel.braun\}@uni-ulm.de}}
\maketitle              
\begin{abstract}
Bounded rationality investigates utility-optimizing decision-makers with limited information-processing power. In particular, information theoretic bounded rationality models formalize resource constraints abstractly in terms of relative Shannon information, namely the Kullback-Leibler Divergence between the agents' prior and posterior policy. Between prior and posterior lies an anytime deliberation process that can be instantiated by sample-based evaluations of the utility function through Markov Chain Monte Carlo (MCMC) optimization. The most simple model assumes a fixed prior and can relate abstract information-theoretic processing costs to the number of sample evaluations. However, more advanced models would also address the question of learning, that is how the prior is adapted over time such that generated prior proposals become more efficient. In this work we investigate generative neural networks as priors that are optimized concurrently with anytime sample-based decision-making processes such as MCMC. We evaluate this approach on toy examples.
 \makeatletter{\renewcommand*{\@makefnmark}{}
\footnotetext{© The Authors (2018) \newline L. Pancioni et al. (Eds.): ANNPR 2018, LNAI 11081, pp. 213–225, 2018. \newline https://doi.org/10.1007/978-3-319-99978-4\_17}\makeatother}

\keywords{Bounded Rationality \and Variational Autoencoder \and Adaptive Priors \and Markov Chain Monte Carlo}
\end{abstract}

\section{Introduction}
Intelligent agents are usually faced with the task of optimizing some utility function $\U$ that is a priori unknown and can only be evaluated sample-wise. We do not restrict ourselves on the form of this function, thus in principle it could be a classification or regression loss, a reward function in a reinforcement learning environment or any other utility function. 
The framework of information-theoretic bounded rationality \cite{Ortega2013,Ortega2015} and related information-theoretic models \cite{Edward2014,Lewis2014,Tishby2011,Todorov2009,Wolpert2006}  provide a formal framework to model agents that behave in a computationally restricted manner by modeling resource constraints through information-theoretic constraints. Such limitations also lead to the emergence of hierarchies and abstractions \cite{Genewein2015}, which can be exploited to reduce computational and search effort. Recently, the main principles have been successfully applied to spiking and artificial neural networks, in particular feedforward-neural network learning problems, where the information-theoretic constraint was mainly employed as some kind of regularization~\cite{Grau-Moya2016,Leibfried2015,Leibfried2017,Peng2017}.
In this work we introduce bounded rational decision-making with adaptive generative neural network priors. We investigate the interaction between anytime sample-based decision-making processes and concurrent improvement of prior policies through learning, where the prior policies are parameterized as Variational Autoencoders \cite{Kingma2013}---a recently proposed generative neural network model.
 
The paper is structured as follows. In Section~\ref{sec:prelim} we discuss the basic concepts of information-theoretic bounded rationality, sampled-based interpretations of bounded rationality in the context of Markov Chain Monte Carlo (MCMC), and the basic concepts of Variational Autoencoders. In Section~\ref{sec:modeling} we present the proposed decision-making model by combining sample-based decision-making with concurrent learning of priors parameterized by Variational Autoencoders. In Section~\ref{sec:empirical} we evaluate the model with toy examples. In Section~\ref{sec:conclusion} we discuss our results.

\section{Methods}
\label{sec:prelim}
\subsection{Bounded Rational Decision Making}
\label{sec:brdm}
The foundational concept in decision-making theory is Maximum Expected Utility \cite{VonNeumann2007}, whereby an agent is modeled as choosing actions such that it maximizes its expected utility
\begin{equation}
\max_{p(a|w)} \sum_w \rho(w) \sum_{a}{p(a|w)\U(w, a)},
\end{equation}
where $a$ is an action from the action space $A$ and $w$ is a world state from the world state space $W$, and $\U(w,a)$ is a utility function. We assume that the world states are distributed according to a known and fixed distribution $\rho(w)$ and that the world sates $w$ are finite and discrete. In the case of a single world state or world state distribution $\rho(w)=\delta(w-w_0)$, the decision-making problem simplifies into a single function optimization problem
$a^* = \argmax_a \U(a)$. In many cases, solving such optimization problems may require an exhaustive search, where simple enumeration is extremely expensive.

A bounded rational decision maker tackles the above decision-making problem by settling on a good enough solution. Finding a bounded optimal policy requires to maximize the utility function while simultaneously remaining within some given constraints. The resulting policy is a conditional probability distribution $p(a|w)$, which essentially consists of choosing an action $a$ given a particular world state $w$. The constraints of limited information-processing resources can be formalized by setting an upper bound on the $\DKL$ (say B bits) that the decision-maker is maximally allowed to spend to transform its prior strategy into a posterior strategy through deliberation. This results in the following constrained optimization problem \cite{Genewein2015}:
\begin{equation}
\max_{p(a|w)} \sum_w \rho(w) \sum_{a}{p(a|w)\U(w, a)}, \text{ s.t. } \DKL(p(a|w)||p(a)) \leq \text{B}.
\end{equation}
This constrained optimization problem can be formulated as an unconstrained problem \cite{Ortega2013}:
\begin{equation}
\max_{p(a|w)} \left( \sum_w \rho(w) \sum_{a}{p(a|w)\U(w, a) - \frac{1}{\beta}\DKL(p(a|w)||p(a))} \right),
\label{eq:bounded}
\end{equation}
where the inverse temperature $\beta \in \mathbb{R}^+$ is a Lagrange multiplier that influences the trade off between expected utility gain and information cost. For $\beta \rightarrow \infty$ the agent behaves perfectly rational and for $\beta \rightarrow 0$ the agent can only act according to the prior policy. The optimal prior policy in this case is given by the marginal $p(a) = \sum_{w \in W}{\rho(w)  p(a|w)}$ \cite{Genewein2015}, in which case the Kullback-Leibler divergence becomes equal to the mutual information, i.e. $\DKL(p(a|w)||p(a))=I(W;A)$. The solution to the optimization problem \eqref{eq:bounded} can be found by iterating the following set of self-consistent equations \cite{Genewein2015}:
\begin{center}
$
\begin{cases}
\begin{array}{rcl}
p(a|w) &=& \frac{1}{Z(w)}p(a) \exp(\beta_1 \U(w,a)) \\
p(a) &=& \sum_w \rho(w) p(a|w), \\
\end{array}
\end{cases}
$
\end{center}
where $Z(w) = \sum_a p(a) \exp(\beta_1 \U(w,a)) $ is normalization factor. Computing such a normalization factor is usually computationally expensive as it involves summing over spaces with high cardinality. We avoid this by Monte Carlo approximation.

\subsection{MCMC as Sample-Based Bounded Rational Decision-Making}
\label{sec:mcmc}
Monte Carlo methods are mostly used to solve two related kinds of problems. One is to generate samples $x$ from a given distribution $q(x)$ and the other is to estimate the expectation of a function. For example, if $g(x)$ is a function for which we need to compute the expectation $\Phi = \E_{q(x)}[g(x)]$ we can draw $N$ samples $\{x_i\}^N_{i=1}$ to obtain the estimate $\hat{\Phi} = \frac{1}{N} \sum_{i=1}^N{g(x_i)}$ \cite{Mackay1998}. Samples can be drawn by employing Markov Chains to simulate stochastic processes. A Markov Chain can be defined by an initial probability $p^0(x)$ and a transition probability $\textbf{T}(x', x)$, which gives the probability of transitioning from state $x$ to $x'$. The probability of being in state $x'$ at the ($t+1)$-th iteration is given by:

\begin{equation}
p^{t+1}(x') = \sum_x{\textbf{T}(x', x)p^t(x)}.
\end{equation}
Such a chain can be used to generate sample proposals from a desired target distribution $q(x)$, if the following prerequisites are met \cite{Mackay1998}. Firstly,  the chain must be ergodic, i.e. the chain must converge  to $q(x)$ independent of the initial distribution $p^0(x)$. Secondly, the desired distribution must be an invariant distribution of the chain. A distribution $q(x)$ is an invariant of $\textbf{T}(x', x)$ if its probability vector is an eigenvector of the transition probability matrix. A sufficient, but not necessary condition to fulfill this requirement is detailed balance, i.e. the probability of going from state $x$ to $x'$ is the same as going from $x'$ to $x$: $q(x)\textbf{T}(x',x) = q(x')\textbf{T}(x,x')$.

An MCMC chain can be viewed as a bounded rational decision-making process for a single context $w$ in the sense that it performs an anytime optimization of a utility function $\U(a)$ with some precision $\gamma$ and that it is initialized with a prior $p(a)$. The target distribution has to be chosen as $q(a)\propto e^{\gamma \U(a)}$ in this case. A decision is made with the last sample when the chain is stopped. The resource corresponds then to the number of steps the chain has taken to evaluate the function $\U(a)$. To find the transition probabilities $\textbf{T}(x',x)$ of the chain, we assume detailed balance and a Metropolis-Hastings scheme $\textbf{T}(x',x)=g(x'|x) A(x'|x)$ such that
\begin{equation}
\label{eq:detailedbalance}
\frac{\textbf{T}(x',x)}{\textbf{T}(x,x')}=\frac{g(x'|x) A(x'|x)}{g(x|x') A(x|x')}=e^{\gamma \left(\U(x')-\U(x)\right)}
\end{equation}
with a proposal distribution $g(x'|x)$ and an acceptance probability $A(x'|x)$.
One common choice that satisfies Equation~\eqref{eq:detailedbalance} is
\begin{equation}
A(x'|x) = \min\left\{1,  \frac{g(x'|x)}{g(x|x')}e^{\gamma \left(\U(x')- \U(x)\right)}\right\}, 
\end{equation}
which can be further simplified when using a symmetric proposal distribution with $g(x'|x)=g(x|x')$, resulting in $A(x'|x) = \min\left\{1,  e^{\gamma \left(\U(x')-\U(x)\right)}\right\}$.

Note that the decision of the chain will in general follow a non-equilibrium distribution, but that we can use the bounded rational optimum as a normative baseline to quantify how efficiently resources are used by analyzing how closely the bounded rational equilibrium is approximated.

\subsection{Representing Prior Strategies with Variational Autoencoders}
\label{sec:vae}
While an anytime optimization process such as MCMC can be regarded as a transformation from prior to posterior, the question remains how to choose the prior. While the prior may be assumed to be fixed, it would be far more efficient if the prior itself were subjected to an optimization process that minimizes the overall information-processing costs. Since in the case of multiple world states $w$ the optimal prior is given by the marginal $p(a)=\sum_w \rho(w)p(a|w)$, we can use the outputs $a$ of the anytime decision-making process to train a generative model of the prior $p(a)$. If the generative model was chosen from a parametric family such as a Gaussian distribution, then training would consist in updating mean and variance of the Gaussian. Choosing such a parametric family imposes restrictions on the shape of the prior, in particular in the continuous domain. Therefore, we investigate non-parametric generative models of the prior, in particular neural network models such as Variational Autoencoders (VAEs).

VAEs were introduced by \cite{Kingma2013} as generative models that use a similar architecture as deterministic autoencoder networks. Their functioning is best understood as variational Bayesian inference in a latent variable model $p(x\vert z,\theta)$ with prior $p(z)$, where $x$ is observable data, and $z$ is the latent variable that explains the data, but that cannot be observed directly.
  The aim is to find a parameter $\hat{\theta}_{ML}$ that maximizes the likelihood of the data $p(x|\theta) = \int p(x\vert  z,\theta)p(z)dz$. Samples from $p(x|\theta)$ can then be generated by first sampling $z$ and then sampling an $x$ from $p(x|z,\theta)$. As the maximum likelihood optimization may prove difficult due to the integral, we may express the likelihood in a different form by assuming a distribution $q(z|x,\eta)$ such that
\begin{eqnarray}
\log p(x|\theta,\eta) &=& \int q(z|x,\eta) \log \frac{p(x|z,\theta)p(z)}{q(z|x,\eta)} \mathop{dz} + \underbrace{\int q(z|x,\eta) \log \frac{q(z|x,\eta)}{p(z|x,\theta)}\mathop{dz}}_{=\DKL(q||p) \geq 0}  \nonumber \\
&\geq& \int q(z|x,\eta) \log \frac{p(x|z,\theta)p(z)}{q(z|x,\eta)} \mathop{dz} \eqqcolon \mathrm{F}(\theta,\eta).
\end{eqnarray}

Assuming that the distribution $q(z|x,\eta)$ is expressive enough to approximate the true posterior $p(z|x,\theta)$ reasonably well, we can neglect the $\DKL$ between the two distributions, and directly optimize the lower bound $\mathrm{F}(\theta,\eta)$ through gradient descent. In VAEs $q(z|x,\eta$ is called the encoder that translates from $x$ to $z$ and $p(x|z,\theta)$ is called the decoder that translates from $z$ to $x$. Both distributions and the prior $p(z)$ are assumed to be Gaussian 

\begin{eqnarray}
p(x|z,\theta) &=& \mathcal{N}\left( x\vert\mu_\theta(z) , \sigma^2 \mathbb{I} \right) \nonumber \\
q(z|x,\eta) &=& \mathcal{N}\left( z\vert\mu_\eta(x) , \Sigma_\eta(x) \right) \nonumber \\
p(z) &=& \mathcal{N}(z|0,\mathbb{I}), \nonumber
\end{eqnarray}
where $\mu_\theta(z)$, $\mu_\eta(x)$ and $\Sigma_\eta(x)$ are non-linear functions implemented by feed-forward neural networks 
and where it is ensured that $\sigma^2 \searrow 0$ and that $\Sigma_\eta(x)$ is a covariance matrix. 

Note that the optimization of the autoencoder itself can also be viewed as a bounded rational choice
  \begin{equation}
  \max_{\theta,\eta}\Bigg(  \mathbb{E}_{q(z|x,\eta)}\left[\log{p(x\vert  z,\theta)}\right]- \DKL\left(q(z\vert  x,\eta)\vert  \vert  p(z)\right) \Bigg),
    \end{equation}
where the expected likelihood is maximized while the encoder distribution $q(z\vert  x,\eta)$ is kept close to the prior $p(z)$.

\section{Modeling Bounded Rationality with Adaptive Neural Network Priors}
\label{sec:modeling}
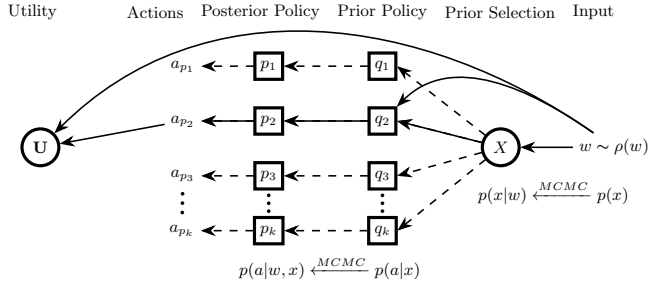
\begin{figure}[t]
\centering
\begin{tikzpicture}[
scale=0.7,
  transform shape, node distance=2cm,
  roundnode/.style={circle, draw=black, very thick, minimum size=7mm},
  squarednode/.style={rectangle, draw=black, very thick, minimum size=5mm},
  arrow/.style = {semithick,-Stealth},
  dotnode/.style={fill,inner sep=0pt,minimum size=2pt,circle} 
]

\node[roundnode] (gate) {$X$};
\node[squarednode, left=1.6cm of gate,yshift=0.5cm] (X2) {$q_2$}; 
\node[squarednode, above=0.5cm of X2]    (X1) {$q_1$};
\node[squarednode, below=0.5cm of X2]    (X3) {$q_3$};
\node[squarednode, below=0.5cm of X3]    (Xk) {$q_k$};

\node[squarednode, left=1.6cm of X2] (PX2) {$p_2$}; 
\node[squarednode, above=0.5cm of PX2]    (PX1) {$p_1$};
\node[squarednode, below=0.5cm of PX2]    (PX3) {$p_3$};
\node[squarednode, below=0.5cm of PX3]    (PXk) {$p_k$};

\node[right=1cm of gate] (x) {$w \thicksim \rho(w)$};
\draw[arrow] (x) -- (gate);
\

\foreach \i/\Yshift in {1/0,2/-3pt,3/-2pt,k/0}
{
   \node[left=1cm of PX\i] (y\i) {$a_{p_\i}$};
   \draw[dashed,arrow] (X\i) -- (PX\i);
   \draw[dashed,arrow] (gate) -- node[""{inner sep=1pt,yshift=\Yshift}]{} (X\i.east);
   \draw[dashed,arrow] (PX\i) -- (y\i);
}
\draw[arrow](PX2) -- (y2);
\draw[arrow](X2) -- (PX2);
\draw[arrow] (gate) -- node[""{inner sep=1pt,yshift=-3pt}]{} (X2.east);
\path (X3) --
  node[dotnode,pos=0.2]{}
  node[dotnode,pos=0.5]{}
  node[dotnode,pos=0.8]{}
  (Xk);
  
\path (y3) --
  node[dotnode,pos=0.2]{}
  node[dotnode,pos=0.5]{}
  node[dotnode,pos=0.8]{}
  (yk);
  
\path (PX3) --
  node[dotnode,pos=0.2]{}
  node[dotnode,pos=0.5]{}
  node[dotnode,pos=0.8]{}
  (PXk);
 
\node[above=.50cm of X1](prior) {Prior Policy}; 
\node[left=.1cm of prior](posterior){Posterior Policy};
\node[left=.1cm of posterior](actions){Actions};
\node[right=.1cm of prior](gating){Prior Selection};
\node[right=.1cm of gating](input) {Input};
\node[left=1.1cm of actions](u){Utility};

\node[below=.15cm of gate, xshift=1cm](selector){$p(x\vert w) \xleftarrow{MCMC} p(x)$};
\node[below=.1cm of Xk, xshift=-1cm](actionselector){$p(a|w,x) \xleftarrow{MCMC} p(a\vert x )$};

\node[roundnode, left=8cm of gate](utility){$\U$};

\draw[arrow](y2) -- (utility);
\draw[arrow](x) to [out=145,in=45] (utility);
\draw[arrow](x) to [out=145,in=45] (X2);
 \end{tikzpicture}
\caption{For each incoming world state $w$ our model samples a prior indexed by $x_i \thicksim p(x\vert  w)$. Each prior $p(a\vert x)$ is represented by a VAE. To arrive at the posterior policy $p(a \vert w,x)$, an anytime MCMC optimization is seeded with $a_0 \thicksim p(a\vert x)$ to generate a sample from $p(a \vert w,x)$. The prior selection policy is also implemented by an MCMC chain and selects agents that have achieved high utility on a particular $w$.}
\label{fig:graphmodel}
\end{figure}
In this section we combine MCMC anytime decision-processes with adaptive autoencoder priors. In the case of a single world state, the combination is straightforward in that each decision selected by the MCMC process is fed as an observable input to an autoencoder. The updated autoencoder is then used as an improved prior to initialize the next MCMC decision. In case of multiple world states, there are two straightforward scenarios. In the first scenario there are as many priors as world states and each of them is updated independently. For each world state we obtain exactly the same solution as in the single world state case. In the second scenario there is only a single prior over actions for all world states. In this case the autoencoder is trained with the decisions by all MCMC chains such that the autoencoder should converge to the optimal rate distortion prior. A third, more interesting scenario occurs when we allow multiple priors, but less than world states---compare Figure~\ref{fig:graphmodel}. This is especially plausible when dealing with continuous world states, but also in the case of large discrete spaces.

\subsection{Decision making with multiple priors} 
Decision-making with multiple priors can be regarded as a multi-agent decision-making problem where several bounded rational decision-makers are combined into a single decision-making process \cite{Genewein2015}. In our case the most suitable arrangement of decision-makers is a two-step process where first each world state is assigned probabilistically to a prior which is then used in the second step to initialize an MCMC chain---compare Figure~\ref{fig:graphmodel}. The output of that chain is then used to train the autoencoder corresponding to the selected prior. As each prior may be responsible for multiple world states, each prior will learn an abstraction that is specialized for this subspace of world states. This two-stage decision-process can be formalized as a bounded rational optimization problem
\begin{equation}
\label{eq:par_mutual}
\max_{p(a|w,x), p(x|w)} \left( \mathbb{E}_{p(a\vert w,x)}[\mathbf{U}(w,a)] - \frac{1}{\beta_1}I(W;X) - \frac{1}{\beta_2}I(W;A|X) \right),
\end{equation}
where $p(x|w)$ is selecting the responsible prior $p(a|x)$ indexed by $x$ for world state $w$. 
The resource parameter for the first selection stage is given by $\beta_1$ and by $\beta_2$ for the second decision made by the MCMC process. The solution of optimization \eqref{eq:par_mutual} is given by the following set of equations:
\begin{equation}
\label{eq:parallelcase}
\begin{cases}
\begin{array}{rcl}
p(x|w) &=& \frac{1}{Z(w)}p(x) \exp(\beta_1 \Delta \F(w,x)) \\[7pt]
p(x) &=& \sum_w \rho(w) p(x|w) \\[7pt]
p(a|w,x) &=& \frac{1}{Z(w,x)} p(a|x) \exp(\beta_2 \mathbf{U}(w,a)) \\[7pt] 
p(a|x) &=& \sum_w p(w|x)p(a|w,x) \\[7pt]
\Delta \F(w,x) &=& \mathbb{E}_{p(a|w,x)}[\mathbf{U}(w,a)] - \frac{1}{\beta_2}\DKL(p(a|w,x)\vert\vert p(a|x)),
\end{array}
\end{cases}
\end{equation}
where $Z(w)$ and $Z(w,x)$ are the normalization factors and $\Delta \F(w,x)$ is the free energy of the action selection stage. The marginal distribution $p(a|x)$ encapsulates an action selection policy consisting of the priors $p(a|w,x)$ weighted by the responsibilities given by the Bayesian posterior $p(w|x)$. Note that the Bayesian posterior is not determined by a given likelihood model, but is the result of the optimization process~\eqref{eq:par_mutual}.

\subsection{Model Architecture}
Equations~\eqref{eq:parallelcase} describe abstractly how a two-step decision process with bounded rational decision-makers should be optimally partitioned.  In this section we propose a sample-based model of a bounded rational decision process that approximately corresponds to  Equations~\eqref{eq:parallelcase} such that the performance of the decision process can be compared against its normative baseline. To translate Equations~\eqref{eq:parallelcase} into a stochastic process we proceed in three steps. First, we implement the priors $p(a|x)$ as Variational Autoencoders. Second, we formulate an MCMC chain that is initialized with a sample from the prior and generates a decision $a\sim p(a|x,w)$. Third, we design an MCMC chain that functions as a selector between the different priors.

\subsubsection{Autoencoder Priors.}
\begin{figure}[t]
\centering
\begin{tikzpicture}[
scale=0.7,
  transform shape, node distance=2cm,
  roundnode/.style={circle, draw=black, very thick, minimum size=7mm},
  squarednode/.style={rectangle, draw=black, very thick, minimum size=5mm},
  arrow/.style = {semithick,-Stealth},
  dotnode/.style={fill,inner sep=0pt,minimum size=2pt,circle} 
]

\node[squarednode] (encoder) {Encoder};
\node[roundnode, below=0.5 of encoder] (input) {$A$};

\node[squarednode, above=0.5cm of encoder, xshift=0.5cm] (mu) {$\mu_\eta(A)$};
\node[squarednode, left=0.1cm of mu] (sigma) {$\Sigma_\eta(A)$};
\draw[arrow] (encoder) -- (mu);
\draw[arrow] (encoder) -- (sigma);
\draw[arrow] (input) -- (encoder);

\node[squarednode, densely dotted, above=0.5cm of sigma, xshift=-1cm] (dkl) {$\DKL(\mu_\eta(A),\Sigma_\eta(A) \vert\vert\mathcal{N}(0, \mathbb{I}))$};
\draw[arrow] (sigma) -- (dkl);
\draw[arrow] (mu) -- (dkl);

\node[squarednode, above=0.5cm of mu, xshift=2cm] (latent) {$z \thicksim \mathcal{N}(\mu_\eta(A), \Sigma_\eta(A))$};
\draw[arrow] (sigma) -- (latent);
\draw[arrow] (mu) -- (latent);

\node[squarednode, above=0.5cm of latent] (decoder) {Decoder};
\draw[arrow] (latent) -- (decoder);

\node[squarednode, above=0.5cm of decoder] (reconstruction) {$a_0 = f_\theta(z)$};
\node[squarednode, densely dotted, left=0.75cm of reconstruction] (reconstructionloss) {$\vert\vert a^* - a_0\vert\vert$};
\draw[arrow] (decoder) -- (reconstruction);
\draw[arrow] (reconstruction) -- (reconstructionloss);

\node[squarednode, right=4cm of encoder, yshift=1cm] (testlatent)  {$z \thicksim \mathcal{N}(0, \mathbb{I})$};
\node[squarednode, above=0.5cm of testlatent] (testdecoder) {Decoder};
\node[squarednode, above=0.5cm of testdecoder] (testreconstruction) {$a_0 = f_\theta(z)$};
\draw[arrow] (testdecoder) -- (testreconstruction);
\draw[arrow] (testlatent) -- (testdecoder);

%
 \end{tikzpicture}
\caption{The encoder translates the observed action into a latent variable $z$, whereas the decoder translates the latent variable $z$ into a proposed action $a$. During training the weights $\eta$ and $\theta$ are adapted to optimize the expected log-likelihood of the observed action samples. After training, the network can generate actions by feeding the decoder network with samples from $\mathcal{N}(z|0,\mathbb{I})$.}
\label{fig:VAE}
\end{figure}
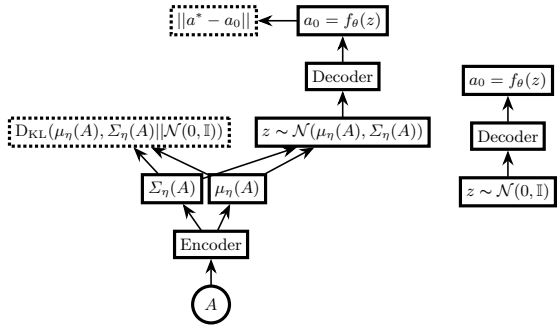
Each prior $p(a|x)$ in Equations~\eqref{eq:parallelcase} is represented by a VAE that learns to generate action samples that mimic the samples given by the MCMC chains---compare Figure~\ref{fig:VAE}. The functions $\mu_\theta(z)$, $\mu_\eta(a)$ and $\Sigma_\eta(a)$ are implemented as feed-forward neural networks with one hidden layer. The units in the hidden layer were all chosen with sigmoid activation function, the output units in the case of the $\mu$-functions were also chosen as sigmoids and for the $\Sigma$-function as ReLU. During training the weights $\eta$ and $\theta$ are adapted to optimize the expected log-likelihood of the action samples that are given by the decisions made by the MCMC chains for all world states that have been assigned to the prior $p(a|x)$. Due to the Gaussian shape of the decoder distribution, optimizing the log-likelihood corresponds to minimizing quadratic loss of the reconstruction error.
After training, the network can generate sample actions itself by feeding the decoder network with samples from $\mathcal{N}(z|0,\mathbb{I})$.

\subsubsection{MCMC Decision-Making.}
To implement the bounded rational decision-maker
$p(a|w,x)$ we obtain an action sample $a\sim p(a|x)$ from the autoencoder prior to initialize an MCMC chain that optimizes the target utility $\mathbf{U}(w,a)$ for the given world state. We run the MCMC chain for $N_{\max}$ steps. In each step we generate a proposal from a Gaussian distribution with $g(a'|a)=\mathcal{N}(a'\vert a,\sigma^2)$ and accept with probability
\begin{equation}
\label{eq:accept2}
A(a'|a) = \min\big\{1, \exp({\gamma(\U(w, a') - \U(w,a))})\big\}.
\end{equation}
Over the course of $N_{\textrm{max}}$ time steps, the precision $\gamma$ is adjusted following an annealing schedule conditioned on the maximum number of steps $N_{\textrm{max}}$. We use an inverse Boltzmann annealing schedule, i.e. $\gamma^{(k)} = \gamma^{0} + \alpha\log(1 + k)$, where $\alpha$ is a tuning parameter. The rationale behind this is that we assume the sampling process to be coarse grained in the beginning and is getting finer during the search. 

\subsubsection{Prior Selection.}
To implement the bounded rational prior selection 
$p(x\vert  w)$ 
through an MCMC process, we first sample an $x$ from the prior $p(x)$ and start an MCMC chain that (approximately) optimizes $\Delta \F(w,x)$ for a given world state $w$ sampled from $\rho(w)$. The prior $p(x)$ is represented by a multinomial and updated by the frequencies of the selected prior indices $x$. The number of steps in the prior selection MCMC chain was kept constant at a value of $N_{\mathrm{max}}^{\textrm{sel}}$ and similarly the precision $\gamma^{\textrm{sel}}$ was annealed over the course of $N_{\mathrm{max}}^{\text{sel}}$ time steps. The target $\Delta \F(w,x)$ comprises a trade-off between expected utility and information resources. However, it cannot be directly evaluated and would require the computation of $\DKL(p(a|x,w)\|p(a|x))$. Here we use number of steps in the downstream MCMC process as a resource measure. As the number of downstream steps was constant, the model selector's choice only depended on the average utility achieved by each decision-maker, which results in the acceptance rule
$$A(x'|x) = \min\left\{1, \exp({\gamma^{\textrm{sel}}(\E_{p(a|w,x)}[\U(w,a)] - \E_{p(a|w,x')}[\U(w,a)]}))\right\}. $$
As the priors are discrete choices the proposal distribution $q(x_{\text{p}}\vert x_\text{p})$ samples globally with $p(x) = \frac{1}{\vert X \vert}$ for all $x$ .
\begin{figure}[t!]
\centering
\includegraphics[width=0.99\textwidth, trim={8.25cm 3cm 9.5cm 5cm}, clip]{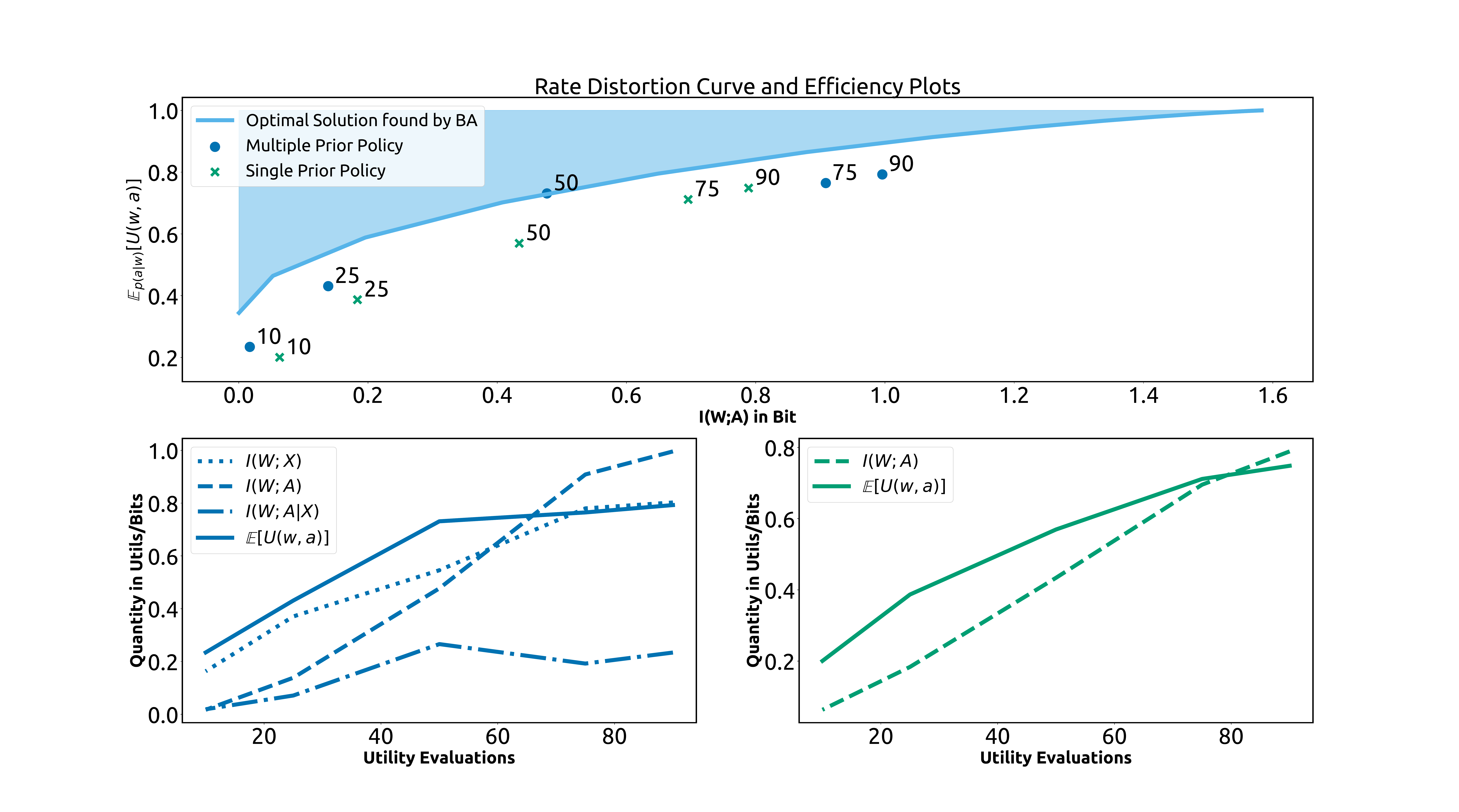}
\caption{Top: The line is given by the Rate Distortion Curve that forms a theoretical efficiency frontier, characterized by the ratio between mutual information and expected utility. Crosses represent single-prior agents and dots multi-prior systems. The labels indicate how many steps were assigned to the second MCMC chain of a total of 100 steps. Bottom: Information processing and expected utility is increasing in the number of utility evaluations, as we expected.}
\label{fig:rd}
\end{figure}
\section{Empirical Results} 
\label{sec:empirical}
To demonstrate our approach we evaluate two scenarios. First, a simple agent, which is equipped with a single prior policy $p_\eta(a)$, as introduced in section \ref{sec:prelim}. In case of a single agent there is no need for a prior selection stage. Second, we evaluated a multi-prior decision-making system and compared the results to the single prior agent. For the mutli-prior agent, we split a fixed number of MCMC steps between the prior selection and the action selection. The task we designed consists of six world states where each world state has a Gaussian utility function in the interval $[0,1]$ with a unique optimum. In both settings, we equipped the Variational Autoencoders with one hidden layer consisting of 16 units with ReLU activations. We implemented the experiments using Keras \cite{Chollet2015}. We show the results in Figure \ref{fig:rd}.

Our results indicate that using MCMC evaluation steps as a surrogate for information processing costs can be interpreted as bounded rational decision-making. In figure \ref{fig:rd} we show the efficiency of several agents with different processing constraints. To compare our results to the theoretical baseline, we discretized the action space into 100 equidistant slices and solved the problem using the algorithm proposed in \cite{Genewein2015} to implement equations \eqref{eq:parallelcase}. Furthermore our results indicate that the multi-prior system generally outperforms the single-prior system in terms of utility.  

\begin{figure}[t!] 
\centering
\includegraphics[width=0.675\textwidth,  trim={7cm 2cm 8.5cm 4cm}, clip]{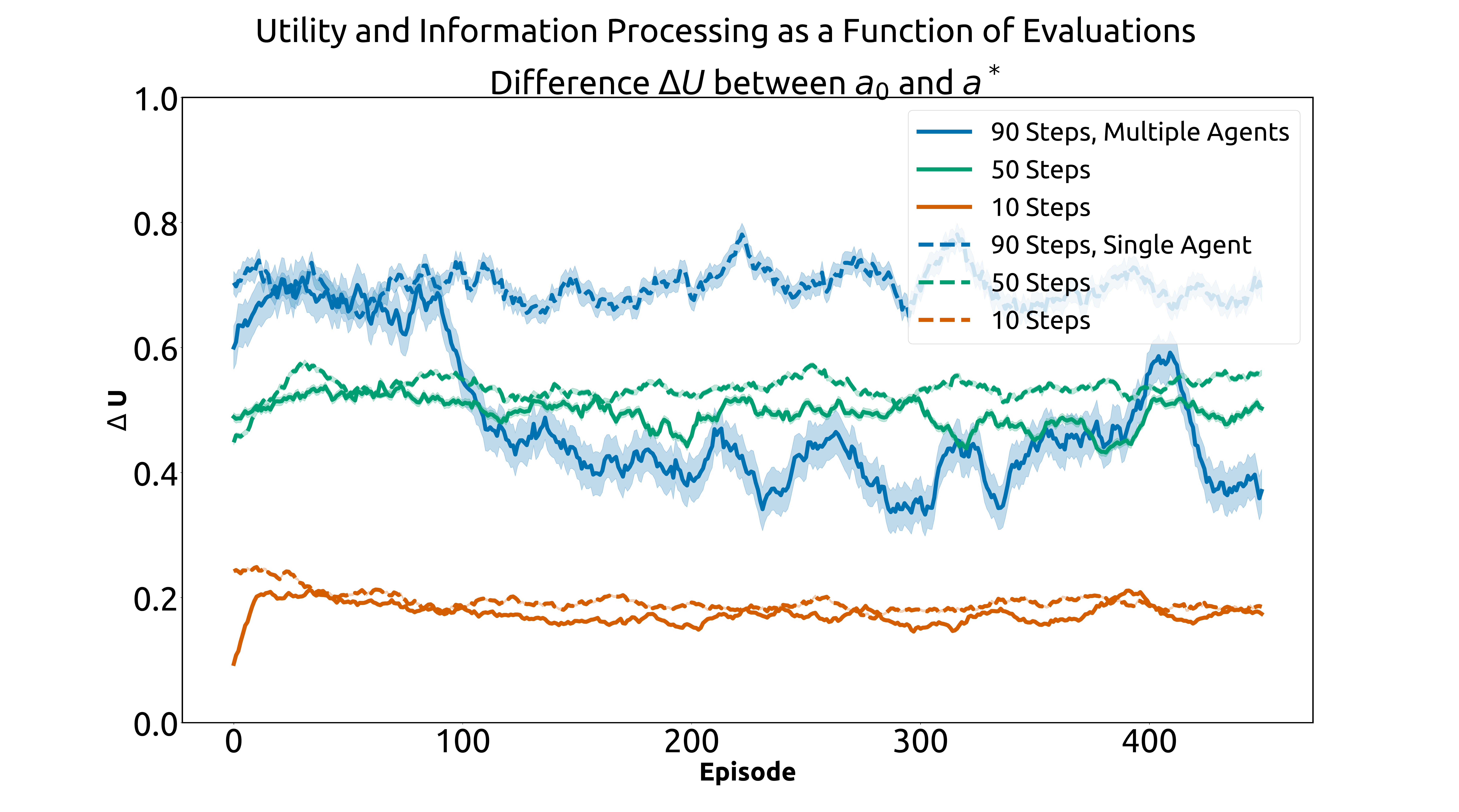}
\caption{Our results indicate that having multiple priors is more beneficial, if more steps are available in total. Note that the stochasticity of our method decreases with the number of allowed steps, as shown by the uncertainty band (transparent regions).}
\label{fig:diffs}
\end{figure}

To illustrate the differences in efficiency between the single prior agent and the multi-prior agents, we plotted in Figure \ref{fig:diffs} utility gained through the second MCMC optimization. For multi-prior agents this is caused by specialized priors which provide initializations to the MCMC chains that are close to the optimal action. In this particular case, $\Delta \U$ does not become zero because we allow only three priors to cover six world states, thus leading to abstraction, i.e. specializing on actions that fit well for the assigned world states. In single-prior agents, the prior is adapting to all world states, thus providing, on average, an initial action that is suboptimal for the requested world state.

\section{Discussion}
\label{sec:conclusion}
In this study we implemented bounded rational decision makers with adaptive priors. We achieved this with Variational Autoencoder priors. The bounded rational decision-making process was implemented by MCMC optimization to find the optimal posterior strategy, thus giving a computationally simple way of generating samples. As the number of steps in the optimization process was constrained,  we could quantify the information processing capabilities of the resulting decision-makers using relative Shannon entropy. Our analysis may have interesting implications, as it provides a normative framework for this kind of combined optimization of adaptive priors and decision-making processes. Prior to our work there have been several attempts to apply the framework of information-theoretic bounded rationality to machine learning tasks \cite{Grau-Moya2016,Leibfried2015,Leibfried2017,Peng2017}. The novelty of our approach is that we design adaptive priors for both the single-step case and the multi-agent case and we demonstrate how to transform information-theoretic constraints into computational constraints in the form of MCMC steps. 

Recently, the combination of Monte Carlo optimization and neural networks has gained increasing popularity. These approaches include both using MCMC processes to find optimal weights in ANNs \cite{Andrieu2000reversible,Freitas2000sequential} and using ANNs as parametrized proposal distributions in MCMC processes \cite{Gu2015Neural,Levy2018generalizing}. While our approach is more similar to the latter, the important difference  is that in such adaptive MCMC approaches there is only a single MCMC chain with a single (adaptive) proposal to optimize a single task, whereas in our case there are multiple adaptive priors to initialize multiple chains with otherwise fixed proposal, which can be used to learn multiple tasks simultaneously. In that sense our work is more related to mixture-of-experts methods and divide-and-conquer paradigms \cite{Ghosh2017,Haruno2001,Yuksel2012}, where we employ a selection policy rather than a blending policy, as we design our model specifically to encourage specialization. In mixture-of-experts models, there are multiple decision-makers that correspond to multiple priors in our case, but experts are typically not modeled as anytime optimization processes.
The possibly most popular combination of neural network learning with Monte Carlo methods was achieved by
AlphaGo \cite{Silver2016}, which beat the leading Go champion by optimizing the strategies provided by value networks and policy networks with Monte Carlo Tree Search, leading to a major breakthrough in reinforcement learning. An important difference here is that the neural network is used to directly approximate the posterior and MCMC is used to improve performance by concentrating on the most promising moves during learning, whereas in our case ANNs are used to represent the prior. Moreover, in our work we assumed the utility function (i.e. the value network) to be given. For future work it would be interesting to investigate how to incorporate learning the utility function into our model to investigate more complex scenarios such as in reinforcement learning.

\subsubsection{Acknowledgement.}
This work was supported by the European Research Council Starting Grant \emph{BRISC}, ERC-STG-2015, Project ID 678082.

\subsubsection{Open Access}
This chapter is licensed under the terms of the Creative Commons Attribution 4.0 International License  \\
(http://creativecommons.org/licenses/by/4.0/), which permits use, sharing, adaptation, distribution and reproduction in any medium or format, as long as you give appropriate credit to the original author(s) and the source, provide a link to the Creative Commons license and indicate if changes were made.

The images or other third party material in this chapter are included in the chapter's Creative Commons license, unless indicated otherwise in a credit line to the material. If material is not included in the chapter's Creative Commons license and your intended use is not permitted by statutory regulation or exceeds the permitted use, you will need to obtain permission directly from the copyright holder.

\bibliographystyle{splncs04}
\bibliography{bibliography.bib}
\end{document}